\documentclass[9pt,technote]{IEEEtran}

\usepackage{wrapfig}
\usepackage{multicol}

\usepackage{amsmath}
\usepackage{amssymb}
\usepackage{graphicx}
\usepackage[noadjust]{cite}
\usepackage{cite}
\usepackage{url}
\usepackage{cite}
\usepackage[section]{placeins}
\usepackage{subfig}
\usepackage[outdir=./]{epstopdf}

\ifCLASSINFOpdf
\else
\fi

\newenvironment{definition}[1][Definition]{\begin{trivlist}
\item[\hskip \labelsep {\bfseries #1}]}{\end{trivlist}}

\newtheorem{theorem}{Theorem}

\newenvironment{proof}{\noindent{\bf Proof:}}{$\hfill \Box$ \vspace{10pt}}
\begin{document}
\title{\LARGE Geometry of locomotion of the generalized Purcell's swimmer : Modelling, controllability and motion primitives}
\author{Sudin~Kadam,
        Ravi~Banavar
\thanks{Sudin Kadam and Ravi Banavar are with Systems and Control Engineering Department, Indian Institute of Technology Bombay, Mumbai,
400076 India e-mail: sudin@sc.iitb.ac.in, banavar@iitb.ac.in}
}

\markboth{Locomotion of the generalized Purcell's swimmer}
{Shell \MakeLowercase{\textit{et al.}}: Bare Demo of IEEEtran.cls for IEEE Journals}

\maketitle

\begin{abstract}
Micro-robotics at low Reynolds number has been a growing area of research over the past decade. We propose and study a generalized 3-link robotic swimmer inspired by the planar Purcell's swimmer. By incorporating out-of-plane motion of the outer limbs, this mechanism generalizes the planar Purcell's swimmer, which has been widely studied in the literature. Such an evolution of the limbs' motion results in the swimmer's base link evolving in a 3-dimensional space. The swimmer's configuration space admits a trivial principal fiber bundle structure, which along with the slender body theory at the low Reynolds number regime, facilitates in obtaining a principal kinematic form of the equations. We derive a coordinate-free expression for the local form of the kinematic connection. A novel approach for local controllability analysis of this 3-dimensional swimmer in the low Reynolds number regime is presented by employing the controllability results of the planar Purcell's swimmer. This is followed by control synthesis using the motion primitives approach. We prove the existence of motion primitives based control sequence for maneuvering the swimmer's base link whose motion evolves on a Lie group. Using the principal fiber bundle structure, an algorithm for point to point reconfiguration of the swimmer is presented. A set of control sequences for translational and rotational maneuvers is then provided along with numerical simulations.
\end{abstract}


\section{Introduction}
Locomotion is one of the most crucial activities for the existential requirements of microbial and animal life. For robots or living organisms it is related to body movements that results in transportation from a physical place to another. There are two major goals in this field - to understand the mechanics of locomotion of existing biological systems, and to devise mobile robotic systems to perform certain desired objectives, possibly by mimicking these biological systems. A commonality in most of the approaches in locomotion is the periodic variation in limbs or shape variables to achieve macroscopic motion. This idea of undulatory limb motion to produce a macroscopic motion is observed in many biological systems from motile microbes to swimming snakes. The idea also appears in a number of engineering settings as presented in \cite{kelly1995geometric}, \cite{morgansen2007geometric}, \cite{shammas2007geometric}.

The interaction with the environment to achieve the motion forms a key to analysis of locomoting systems. Most of our intuition about locomotion originates from inertia-dominant systems, prevalent in larger animals. As opposed to such systems, micro-organisms resort to a creeping motion. The usual mechanism for swimming in water for larger animals involves obtaining a forward momentum from the surrounding fluid due to some periodic body motion. The effect of inertia is that the displacement gained in the first half period of a cyclic motion is not cancelled out by that of the second half period \cite{childress1981mechanics}. Such a mechanism, however, does not work in the microscopic world of biological objects where the inertia is negligible and viscosity dominates the motion. This effect is observed at very low Reynolds numbers, a nondimensional ratio of the inertial to viscous forces acting on the swimmer body.  A vast majority of living organisms are found to perform motion at microscopic scales, where the viscous forces strongly dominate the motion at low Reynold's number. The Reynolds number in such regimes is of the order of $10^{-4}$. To get a comparative idea of this number, the Reynolds number for a human swimming in water is of the order of $10^4$, for a Goldfish swimming in water it is of the order of $10^2$, and that for a human trying to swim in honey is of the order of $10^{-3}$ \cite{cohen2010swimming}, \cite{najafi2004simple}.

\begin{figure}[htb!]
\minipage{0.227\textwidth}
\begin{center}
  \includegraphics[width=0.75\linewidth]{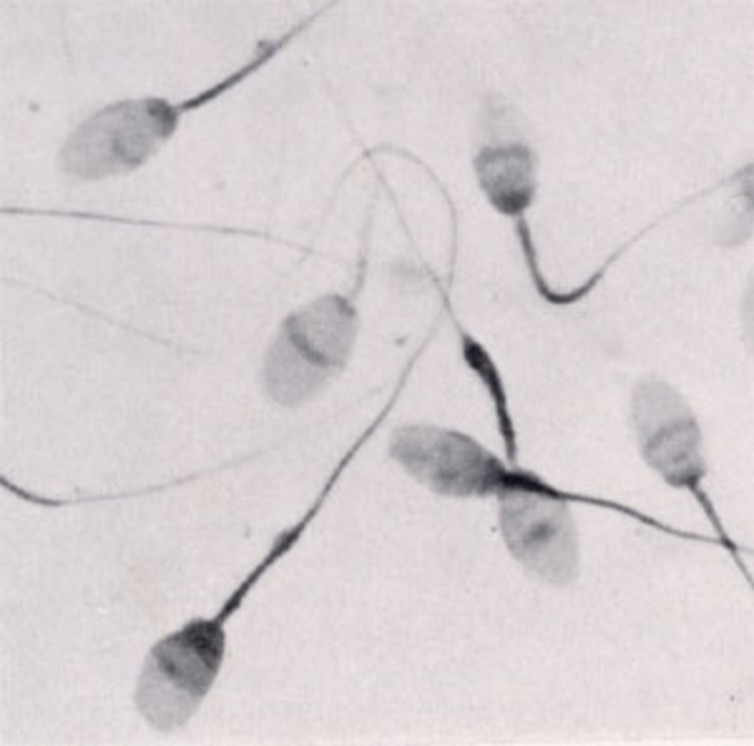}
  \caption{Sperm cells \cite{sperm}}\label{fig:sperm1}
\end{center}
\endminipage\hfill
\minipage{0.277\textwidth}%
\begin{center}
  \includegraphics[width=0.95\linewidth]{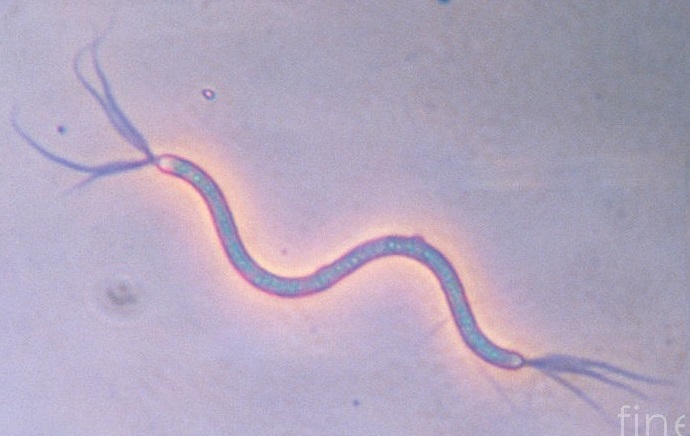}
  \caption{Spirillum bacteria \cite{spirrilum}}\label{fig:spirillum_volutans}
  \end{center}
\endminipage
\end{figure}

\begin{figure}
  \begin{center}
    \includegraphics[width=0.48\textwidth]{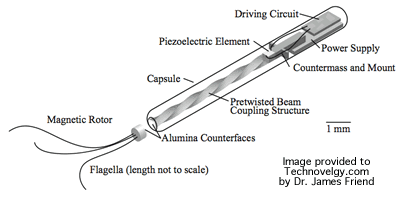}
    \caption{Flagellar microbot \cite{Technovelgy}}\label{fig:micromotor}
  \end{center}
\end{figure}

There has been a lot of research and a growing interest in exploring new and efficient ways to generate motion at these micro scales, see \cite{becker2003self}, \cite{qiu2014swimming}. Understanding the physics of locomotion in microbes provides many insights. Organisms such as paramecia are covered with cilia that are used much like little oars to synchronously row through water. Motile bacteria use one or more flagella, which are long helical filaments that extend from the cell membrane and act as a propeller. Escherichia Coli is a bacteria commonly found in the lower intestines of humans which, for example, uses such a mechanism to swim at a speed of $35$ diameters per second \cite{berg2008coli}. Figures \ref{fig:sperm1} and \ref{fig:spirillum_volutans} show a few such microbes with flagella. In the field of engineering, microbotics (or microrobotics) is one of the recently evolving fields in mobile robots with characteristic dimensions to the scale of a micron. Figure \ref{fig:micromotor} shows an example of a flagellar microswimmer developed at Monash University. Cell or microbial locomotion being an essential part of biological systems, understanding the mechanism of locomotion is also of a great interest to biologists community. A better understanding of the mechanism of swimming can lead to many useful applications in  several fields such as targeted drug delivery, non-evasive surgery, micro-machining and nano technology.

The slender-body theory and the resistive force theory form much of the basis for modelling of the fluid forces acting on the systems in this regime \cite{cox1970motion}. These are used to obtain a relationship between the velocity of the body at each point along its length and the force per unit length experienced by the body at that point. Moreover, the resistive force theory allows us to treat the forces acting on the individual limbs independent of the motion of the rest of the motion of the other parts of the locomoting body \cite{hatton2013geometric}. These features greatly simplify the fluid mechanistic modelling.

From the modelling and controls perspective, the locomotion of articulated mechanical systems is often complex. Classical mechanics tends to become intractable for such systems, and tends not to reveal the topological structure of the systems' configurations, velocities and other quantities. Geometric mechanics and control theory aid considerably in the analysis of robotic and animal locomotion, where the treatment of the configuration space as a differentiable manifold allows one to extend the calculus to topological spaces which are not Euclidean \cite{bloch2003nonholonomic}, \cite{bullo2004geometric}, \cite{holm2009geometric}, \cite{ostrowski1998geometric}. For a large class of locomotion systems, including underwater vehicles, fishlike swimming, flapping winged vehicles, spacecraft with rotors and wheeled or legged robots, it is possible to model the motion using the mathematical structure of a connection on a principal bundle, see  \cite{ostrowski1998geometric}, \cite{cabrera2008base} for examples.

In his famous lecture on "Life at Low Reynolds Number \cite{purcell1977life}", E. M. Purcell presented the simplest swimmer that can effectively propel itself at low Reynolds numbers. This swimmer can be considered as a simplified flagellum made of three slender rods articulated at two hinges. This mechanism gave rise to a lot of research in modelling, control, optimal gait design etc. of this swimmer, see \cite{avron2008geometric}, \cite{burton2013dynamics}, \cite{melli2006motion}, \cite{najafi2004simple}, \cite{passov2012dynamics} and the references therein. \cite{hatton2013geometric} analyzes its locomotion problem in a geometric framework, again for the planar case, which uses the low Reynold's number regime and slenderness of the links in mechanism to get a kinematic form of equations. The Cox theory at low Reynold's number for slender bodies is not restricted to planar motion. We extend existing work to a more general, more challenging and more practical 3-dimensional locomotion problem by using tools from geometric mechanics.

\subsection{Contribution:}
The existing literature presents various aspects of modelling and control of the planar Purcell's swimmer. Since microbial and biomimetic motion is certainly not restricted to a plane and evolves in 3 dimensional space, there is a need to study and analyze a 3D model. To our knowledge, this article presents the first study of a 3 link swimming mechanism which performs generic 3-dimensional motion. Furthermore, by adopting a geometric approach, we identify certain structures in the configuration space which leads to an insightful model for such a complex system. Such a geometric approach for the planar Purcell’s swimmer is given in \cite{hatton2008connection}. We derive a coordinate free expression of the local connection form which avoids the cumbersome notation of local parametrizations right from the onset in modelling. We then present a local, weak-controllability analysis at a particular configuration which characterizes the allowable group motions of the swimmer. This is followed by control synthesis using the motion primitives approach. A set of control sequences is synthesized to generate motions along the basis of the Lie algebra of the structure group of the swimmer.

\subsection{Organization of paper:}
The paper is organized as follows. In the next section we present the geometry that typical locomotion systems admit, followed by that of the proposed generalized Purcell's swimmer. In section $3$, we derive a coordinate-free expression for the principal kinematic form of the swimmer using the Cox and resistive force theories. Section $4$ shows the local weak-controllability analysis of the swimmer in its collinear configuration. In the section $5$, we prove the existence of a motion primitive based control law, resulting in a control sequence for certain group motions along with simulation results.

\section{Construction and the geometry of configuration space}
\begin{figure}
  \begin{center}
\includegraphics[scale=0.55]{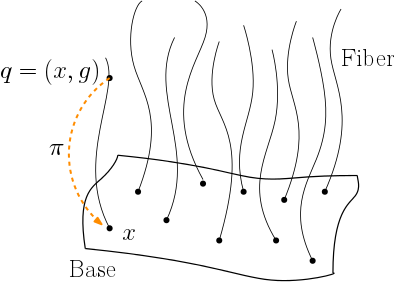}
\caption{Fiber Bundle}
\label{fiber_bundle}
  \end{center}
\end{figure}

While studying the problem of locomotion using limb motion or shape change, the geometry of the configuration space, which is a differential manifold requires attention for elegant and insightful solutions. The configuration space is usually written as the product of two manifolds. One part is the base manifold $M$ which describes the configuration of the internal shape variables of the mechanism. The other part depicting the macro-position of the locomoting body is a Lie group $G$ and represents gross displacement of the body. The total configuration space of the robot $Q$ then naturally appears as a product $G \times M$. Such systems follow the topology of a trivial principal fiber bundle, see \cite{kobayashi1963foundations}. Figure \ref{fiber_bundle} shows an explanatory figure of a fiber bundle. With such a separation of the configruation space, locomotion is readily seen as the means by which changes in shape affect the macro position. We refer to \cite{kelly1995geometric}, \cite{bloch1996nonholonomic} for a detailed explanation on the topology of locomoting systems.

\subsection{Generalized Purcell's swimmer}
The original form of the Purcell's swimmer has three links moving in a plane, the outer links are actuated through a hinge joint with the central or base link such that the mechanism always performs motion in a plane. The details of construction, configuration manifold and the kinematic modelling of this swimmer can be referred from \cite{hatton2013geometric}. Fig. \ref{fig:3d_coord_free} shows a schematic of the proposed generalization. We replace the two hinge joints by ball joints, thus allowing out-of-plane motion of the mechanism through yaw, pitch and roll motions of the 2 outer limbs. The middle or base link manoeuvres in the Special Euclidean group $SE(3)$. The translational position of the central (base) link denotes the position of its center and is an element of $\mathbb{R}^3$, and its rotational position is an element of the Special Orthogonal group of matrices $SO(3)$. Each link in our swimmer is modelled as a rigid slender body of length $2L$. The orientation of each of the two outer links also evolve on $SO(3)$. Hence the total configuration space of the proposed swimmer is given by $Q\:=\: SO(3) \times SO(3) \times SE(3)$

\begin{definition}:
\textit{For $Q$ a configuration manifold and $G$ a Lie group, a trivial principal fiber bundle with base $M$ and structure group $G$ is a manifold $Q = M \times G$ with a free left action of $G$ on $Q$ given by left translation in the group variable: $\phi_h(x,g) = (x,hg)$ for $x \in M$ and $g \in G$} \cite{kelly1995geometric}
\end{definition}
The structure group in our case is $G = SE(3)$. and the shape space $M$ is $SO(3) \times SO(3)$. Since all the points $q \in Q$ are represented by $(x, g)$ with $x \in M$ and $g \in G$,  $Q$ has global product structure of the form $M \times G$. Moreover, $SE(3)$ acts via the left action as a matrix multiplication, and has a single identity element, which is a $4 \times 4$ identity matrix. The left action of the group, defined by $\Phi_h : (x,g) \in Q \longrightarrow (x,hg)$ is hence seen to be free, for $x \in M$  and $\: h,g \in G$. The configuration space of the basic Purcell's swimmer satisfies the trivial principal fiber bundle structure.

\section{Kinematic Model of the Swimmer}
\subsection{Fluid Forces}
Due to the assumption of low Reynold's number regime, the hydrodynamics of the system is governed by Stokes equations, which are the simplified Navier-Stokes equations  \cite{hatton2011geometric}. For a slender body at low Reynold's number, Cox theory gives a linear dependence of forces on velocities, see \cite{cox1970motion} . This implies that for a slender link of length $l$, radius $a$, fluid viscosity $\mu$ at speeds $u_long$ and $u_lat$ in longitudinal and lateral directions, respectively, the forces acting are given by
\begin{align*}
F_{long} &= \frac{2\pi\mu l u_{long}}{ln(\frac{l}{a})} = k_Tu_{long} \:\:\:\:\:\:  \text{(Along the link length)} \\
F_{lat} &= \frac{4\pi\mu l u_{lat}}{ln(\frac{l}{a})} = 2k_Tu_{lat} \qquad \text{(Perpendicular to the link length)}
\end{align*}
where $k_T$ is the differential viscous drag constant corresponding to translational motion. We model each of the three links of the swimmer as slender members leading to the form of fluid forces' according to Cox theory. We regard the flows around each link as independent of the other links' motions according to resistive force theory \cite{tam2007optimal}, thus eliminating the hydrodynamic coupling between the links. We denote by $\xi_{i,x}$ the translational velocity along the longitudinal axis, and by $\xi_{i,y}$, $\xi_{i,z}$ the translational velocities along the 2 lateral axes of the i'th link. Similarly, $\xi_{i,\omega_x}, \xi_{i,\omega_y}$ and $\xi_{i,\omega_z}$ denote angular velocities about the respective 3 body axes of the i'th link. The total forces and moments acting on the i'th link of length $2L$, in its own frame take the following form
\begin{align*}
F_{i,x}\: &=\:\int_{-L}^L \frac{1}{2}k_T\xi_{i,x} dL \:=\: k_T \xi_{i,x} L \\
F_{i,y}\: &=\:\int_{-L}^L k_T\xi_{i,y} dl \:=\: 2 k_T \xi_{i,y} L \\
F_{i,z}\: &=\:\int_{-L}^L k_T\xi_{i,z} dl \:=\: 2 k_T \xi_{i,z} L
\end{align*}
The moments about the center of the link due to rotation about an axis transverse to link is found by taking the lateral drag forces as linearly varying along the link according to its angular velocity.
\begin{figure}[h!]
\centering
\includegraphics[scale=.37]{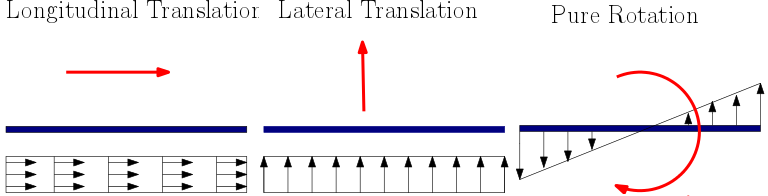}
\caption{Force and moments on a slender member}
\label{fig:1 3Link arbit}
\end{figure}
Note that the Cox theory is restricted to forces and moment acting on the body performing only translational motion. Hence, to account for the torque acting on a slender link due to spinning about its own axis, we define  $K_R$ as the resistive torque per unit length per unit spin speed. Thus, we have total moments as
\begin{align*}
M_{i}\:&=\:\int_{-L}^L [k_R\xi_{i,\omega_x},\: k_Tl^2\xi_{i,\omega_y},\: k_Tl^2\xi_{i,\omega_z}]^T dl \\ & =\: [k_R L\xi_{i,\omega_x},\: \frac{2}{3}k_T L^3\xi_{i,\omega_y},\: \frac{2}{3}k_T L^3\xi_{i,\omega_z}]^T
\end{align*}
Thus, the total forces and moments on each link can be written as linear equation in its body velocities -
\begin{equation}\nonumber
F_i\:=\:\begin{bmatrix}
k_TL & 0 & 0 & 0 & 0 & 0 \\
0 & 2k_TL & 0 & 0 & 0 & 0 \\
0 & 0 & 2k_TL & 0 & 0 & 0 \\
0 & 0 & 0 & k_RL & 0 & 0 \\
0 & 0 & 0 & 0 & \frac{2}{3} kL_T^3 & 0 \\
0 & 0 & 0 & 0 & 0 & \frac{2}{3} kL_T^3
\end{bmatrix}\begin{pmatrix}
    \xi_{i,x} \\
    \xi_{i,y} \\
    \xi_{i,z}	\\
    \xi_{i,\omega_x} \\
    \xi_{i,\omega_y} \\
    \xi_{i,\omega_z} \\
\end{pmatrix}
\end{equation}
which, for the $i^{th}$ link can be simply written as
\begin{equation}\label{Force_ith_link}
F_i\:=\:H_i\xi_i
\end{equation}
\subsection{Coordinate frames and transformations}
\begin{figure}[!htbp]
\centering
\includegraphics[scale=.52]{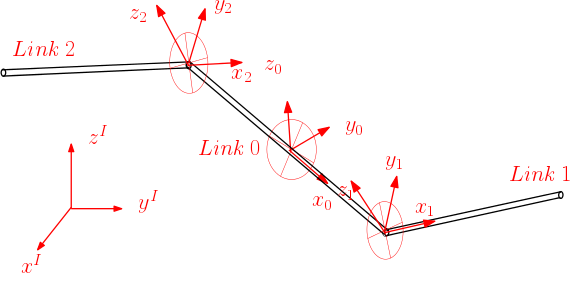}
\caption{Generalized Purcell's swimmer configuration}
\label{fig:3d_coord_free}
\end{figure}

Fig \ref{fig:3d_coord_free} shows an arbitrary position of the system along with 3 coordinate frames corresponding to each link. The frame corresponding to base link has its origin at its geometric center, whereas the frames corresponding to the outer links are at their respective joints with base link.  The orientation of the outer links is represented using an element of the Special Orthogonal group of $3 \times 3$ matrices. Thus $R_1 ,R_2 \in SO(3)$ define the coordinates of the shape space $M$. The reference configuration is the one in which all the 3 coordinate frames are aligned to each other, i.e. $R_1$ and $R_2$ both are the identity matrices.

We write the kinematics of the i'th link in terms of the velocity of the base link $\xi_0$ and the angular velocities of the 2 outer links. The velocity of i'th link in its own body frame is an element of the Lie algebra $\mathfrak{g}$, which in our case is $se(3)$. It is represented as a generalized velocity vector by an element of $\mathbb{R}^6$ as $\xi_i\:=\: \begin{bmatrix}
    \xi_{i,x}, \: \xi_{i,y},\: \xi_{i,z},\: \xi_{i,\omega_x},\: \xi_{i,\omega_y},\: \xi_{i,\omega_z} \end{bmatrix}^T$.

\subsection{Construction of the connection form}
A mechanical system having a principal fiber bundle structure and group symmetry in Lagrangian can be shown to satisfy the general reconstruction equation
\begin{equation}\label{reconstruction_eqn}
\xi = -A(r)\dot{r} + \mathbb{I}^{-1}p
\end{equation}
where, for a shape manifold $M$, its tangent space at a point $r \in M$ is denoted by $T_rM$, the local connection form is $A(r) : T_rM \longrightarrow \mathfrak{g}$, $\mathbb{I} : \mathfrak{g} \longrightarrow \mathfrak{g}^*$ is the locked inertia tensor and $p \in \mathfrak{g}^*$ is the generalized momentum of the system, see \cite{bloch2003nonholonomic}, \cite{shammas2007geometric} for details on kinematic systems. For swimming at low Reynolds number, the strong dominance of viscous forces means that the momentum terms drops out.  The reconstruction equation takes the following kinematic form
\begin{equation}\label{kin_reconstruction}
\xi = -A(r)\dot{r}
\end{equation}
In the subsequent part of the paper, we obtain a model of the 3 dimensional Purcell's swimmer in this kinematic form. In order to derive the force on each link, from equation  \ref{Force_ith_link}, we see that we need to find the body velocity of each link in terms of the body velocity of the base link and the shape velocities $\omega_1 = \begin{bmatrix}
\xi_{1,\omega_x},\: \xi_{1,\omega_y},\: \xi_{1,\omega_z} \end{bmatrix}^T$ and $\omega_2 = \begin{bmatrix}
\xi_{2,\omega_x},\: \xi_{2,\omega_y},\: \xi_{2,\omega_z} \end{bmatrix}^T$, which are the angular velocities of links $1$ and $2$ respectively in their own frames.  $\omega_1,\: \omega_2$ also constitute the inputs to our kinematic system. The relationship between body and shape velocities is obtained for link 1 as

\begin{equation}\label{xi1}
\begin{split}
\xi_1\:&= \left[
\begin{array}{cccc}
R_1 &	-\begin{bmatrix}
l \\
0 \\
0
\end{bmatrix}^\times &  0_{3\times3} &  0_{3\times3} \\
0  & R_1  & e_{3\times3} &  0_{3\times3}
\end{array}\right]
\begin{pmatrix}
    \xi_0 \\
    \omega_1 \\
    \omega_2
\end{pmatrix} \: = \: B_1 \begin{pmatrix}
    \xi_0 \\
    \omega_1 \\
    \omega_2
\end{pmatrix}
\end{split}
\end{equation}
where $0_{3\times3}$ and $e_{3\times3}$ are zero and identity matrices of size ${3\times3}$, respectively. The hat map $(.)^{\times}$ is defined in appendix \ref{A}. A similar expression can be derived for the second link as

\begin{equation}\label{xi2}
\begin{split}
\xi_2\:&= \left[
\begin{array}{cccc}
R_2 &	\begin{bmatrix}
l \\
0 \\
0
\end{bmatrix}^\times &  0_{3\times3} &  0_{3\times3} \\
0  & R_2  &  0_{3\times3} & e_{3\times3}
\end{array}\right]
\begin{pmatrix}
    \xi_0 \\
    \omega_1 \\
    \omega_2
\end{pmatrix} \: = \: B_2 \begin{pmatrix}
    \xi_0 \\
    \omega_1 \\
    \omega_2
\end{pmatrix}
\end{split}
\end{equation}
This leads us to the following form of the velocity of each of links 0, 1 and 2
\begin{equation}\label{B_matrix}
\xi_0=\begin{pmatrix}
    \xi_{0,x} \\
    \xi_{0,y} \\
    \xi_{0,z}	\\
    \xi_{0,\omega_x} \\
    \xi_{0,\omega_y} \\
    \xi_{0,\omega_z}
\end{pmatrix}, \qquad \xi_1=B_1\begin{pmatrix}
    \xi_0 \\
    \omega_1 \\
    \omega_2
\end{pmatrix}, \qquad \xi_2=B_2\begin{pmatrix}
    \xi_0 \\
    \omega_1 \\
    \omega_2
\end{pmatrix}
\end{equation}
Next, from fluid force equations (\ref{Force_ith_link}), we see that the force on each link is linearly dependent on the velocity of the link in its own frame. Hence the force on each of the 3 links is

\begin{equation}
F_0\:=\:H_0\xi_0, \:\: F_1\:=\:H_1B_1 \begin{pmatrix}
    \xi_0 \\
    \omega_1 \\
    \omega_2 \end{pmatrix}, \:\: F_2\:=\:H_2B_2 \begin{pmatrix}
    \xi_0 \\
    \omega_1 \\
    \omega_2
    \end{pmatrix}
\end{equation}
The summation of all the forces gives us the resultant force acting on the system. But we note that equation (\ref{Force_ith_link}) gives us the force with respect to the frame of the respective link. Hence before summing up, we transform the forces to the frame associated with base link as follows
\begin{equation}
F_{net}^0\:=\:F_0 + T_1^0F_1 + T_2^0F_2
\end{equation}
where the transformation matrices corresponding to outer links are
\begin{equation}\label{Transformation_matrices}
T^0_1=\begin{bmatrix}
R_1^T & 0\\
\begin{bmatrix}
l\\
0\\
0
\end{bmatrix}^\times R_1^T  & R_1^T
\end{bmatrix}, \qquad
T^0_2=\begin{bmatrix}
R_2^T & 0\\
\begin{bmatrix}
-l\\
0\\
0
\end{bmatrix}^\times R_2^T  & R_2^T
\end{bmatrix}
\end{equation}
We substitute the forces on each link using equations \ref{Force_ith_link} and \ref{B_matrix} and then split these equations by writing matrices in terms of their block format as
\begin{align*}
F&=H_0\xi_0 + T^0_1 H_1B_1 \begin{pmatrix}
    \xi_0 \\
    \omega_1 \end{pmatrix} + T_2^0H_2B_2 \begin{pmatrix}
    \xi_0 \\
    \omega_2 \end{pmatrix} \\
   &= H_0\xi_0 + [[T^0_1 H_1B_1]_{6\times6} \:\: [T^0_1 H_1B_1]_{6\times3}] \begin{pmatrix}
    \xi_0 \\
    \omega_1 \end{pmatrix} \\ 
    & \qquad \qquad + [[T^0_2H_2B_2]_{6\times6} \:\: [T^0_2H_2B_2]_{6\times3}] \begin{pmatrix}
    \xi_0 \\
    \omega_2 \end{pmatrix}
\end{align*}
The consequence of being at low Reynolds number is that the net forces and moments on an isolated system should be zero, which leads to following equation
\begin{align*}
0\:&=\:H_0\xi_0 + [T^0_1 H_1 B_1]_{6\times6}\:\xi_0 +[T^0_1 H_1 B_1]_{6\times3}\: \omega_1 + \\ 
& \qquad  [T^0_2 H_2 B_2]_{6\times6}\:\xi_0 + [T^0_2 H_2 B_2]_{6\times3}\: \omega_2
\end{align*}
Finally, by rearranging the terms we get reconstruction equation as
\begin{equation}\label{Final_kinematic_model}
\xi_0 = -P^{-1}Q \:\dot{r}
\end{equation}
where $\dot{r} = [\omega_1,\: \omega_2 ]^T$ are the shape velocities and $P,\: Q$ matrices are given by
\begin{align*}
P &=  [H_0+[T^0_1H_1B_1]+[T^0_2H_2B_2]]_{6\times6}, \\ 
Q &=  [[T^0_1 H_1 B_1]_{6\times3}\:\:\:\: [T^0_2 H_2 B_2]_{6\times3}]_{6\times6}
\end{align*}
This is the desired model in a principal kinematic form, $\xi = -A(r)\dot{r} $, where $A(r)$ is the local connection form defined at each $r = (R_1, R_2) \in M = SO(3) \times SO(3)$ and the shape velocity $\dot{r} = [\omega_1,\: \omega_2 ]^T \in T_rM$. For the proposed 3 dimensional Purcell's swimmer, the local connection form $A(r)$ is a $6 \times 6$ matrix, which depends on the lengths of the limbs, viscous drag coefficients $k_T$, $k_R$ and the shape of the mechanism $r$. In our example the Lie group is the Special Euclidean group $SE(3)$, and $\xi=[v_x, v_y, v_z, \omega_x, \omega_y, \omega_z]^T$ belongs to its tangent space at the identity. The connection form and the other notions mentioned here have roots in geometric mechanics, see \cite{bloch2003nonholonomic}, \cite{holm2009geometric} for details.

\section{Local controllability analysis}
The overall kinematics of the swimmer can be written in the form of a driftless linear-in-control system as
\begin{equation}\label{pure_kinematic2}
\begin{bmatrix}
\dot{r} \\
\xi_0
\end{bmatrix} = \begin{bmatrix}
e_{6\times6} \\
-A(r)
\end{bmatrix} \dot{r}
\end{equation}
where, $e_{6 \times 6}$ is a $6 \times 6$ identity matrix and $\mathbb{R}^6 \ni \dot{r} = [\omega_1, \omega_2]^T \in T_rM$ is the control input. This is a principal kinematic form of equations, which we write in terms of the control vector fields $g_1^1 \cdots g_3^1, g_1^2 \cdots g_3^2 $ as follows
\begin{equation}\label{pure_kinematic_CVFs}
\begin{bmatrix}
\dot{r} \\
\xi_0
\end{bmatrix} = \sum_{i=1}^3 g_i^1 \omega_1^i  + \sum_{i=1}^3 g_i^2 \omega_2^i
\end{equation}
where the index $i$ in $\omega^i_j$ corresponds to the joint velocity components along the $i^{th}$ axis of $j^{th}$ limb's frame. The use of Chow's theorem to comment on the local controllability of this system requires us to check the Lie algebra rank condition \cite{jurdjevic1997geometric} of the control vector fields $g_i^j$'s. We tried to do this for the 3D Purcell's swimmer for a configuration where all the $3$ links are collinear with each other. But computing the Lie algebra of the control vector fields requires calculation of successive Lie brackets, which requires the explicit form of these vector fields. Due to the matrix inversion step involved (see equation \ref{Final_kinematic_model}), the expression for these vector fields becomes extremely bulky, which, consequently, makes Lie bracket calculation computationally highly intensive. Moreover, in the midst of such bulky terms, the physical insight into the system gets lost. Thus, we have used an alternate and indirect approach based on the controllability results of the planar Purcell's swimmer to comment on the local weak controllability of the proposed 3D swimmer in the collinear configuration. This configuration corresponds to the identity point $e$ of the base space $M = SO(3) \times SO(3)$.
\subsection{Weak controllability of the 3D Purcell's swimmer}
\begin{figure}[htb!]
\begin{center}
  \includegraphics[width=0.27\textwidth]{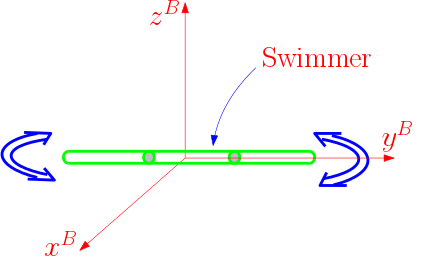}
\caption{Motion in XY plane}
\label{Limb motion in XY plane}
\end{center}
\end{figure}

\begin{figure}[htb!]
\begin{center}
\includegraphics[width=0.27\textwidth]{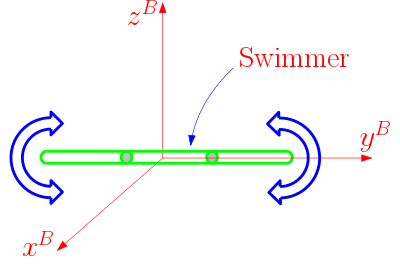}
\caption{Motion in YZ plane}
\label{Limb motion in YZ plane}
  \end{center}
\end{figure}

The principal fiber bundle structure in the controllability analysis naturally gives rise to a finer notion of weak controllability, described in appendix \ref{B}. This idea is also of practical relevance since often just reaching the desired group component without strict requirement on shape of the system is sufficient. We make use of the controllabilty results from \cite{giraldi2013controllability}, \cite{kadamgeometric}, which state that the planar version of the Purcell's swimmer is strongly controllable at this collinear configuration. We decompose the limb motion of the 3D swimmer into that of 2 planar swimmers as shown in the figures \ref{Limb motion in XY plane}, \ref{Limb motion in YZ plane}, and use the result that in each of these planes the planar swimmer's motion on $SE(2)$ group is locally strongly controllable.

\begin{theorem}
The 3D Purcell's swimmer is locally weakly-controllable in its collinear configuration
\end{theorem}
\begin{proof}
Since the 3D swimmer's limbs are actuated in both the $x^B-y^B$ and $y^B-z^B$ planes, the planar Purcell's swimmer's controllability results imply that,
\begin{itemize}
\item For the case of limb actuation in $x^B-y^B$ plane, the swimmer's translational motion is locally controllable in the $x^B$ and $y^B$ axis directions, and the rotational motion about the $z^B$ axis is controllable. Hence, the achievable motion is in the Lie algebra of $SE(2)$, given as
\begin{equation}
\mathfrak{g}_1=span\{ \xi_{0,x}, \xi_{0,y}, \xi_{0,\omega_z}\}
\end{equation}
\item For the case of limb actuation in $y^B-z^B$ plane, the swimmer's translational motion is locally controllable in the $y^B$ and $z^B$ axis directions, and the rotational motion about the X axis is controllable. i.e. the achievable motion is in the Lie algebra of $SE(2)$, given as
\begin{equation}
\mathfrak{g}_2=span\{ \xi_{0,y}, \xi_{0,z}, \xi_{0,\omega_x}\}
\end{equation}
\item In the collinear configuration of the swimmer, just the rotation of the the outer links in mutually opposite directions about their own axis leads to the pure rotation of the swimmer's base link about its $x$ axis
\begin{equation}
\mathfrak{g}_3=span\{\xi_{0,\omega_y}\}
\end{equation}
\item Thus, $\mathfrak{g}_1, \mathfrak{g}_2$ and $\mathfrak{g}_3$ together span the Lie algebra of $SE(3)$ completely.
\begin{equation}
\mathfrak{g}= se(3) = \mathfrak{g}_1 + \mathfrak{g}_2 + \mathfrak{g}_3
\end{equation}
\end{itemize}
Since the Lie algebra of the structure group $SE(3)$ is spanned by the Lie algebra generated by the control vector fields, at this collinear configuration the 3D swimmer is locally weakly controllable.
\end{proof}

\subsection{Point-to-point reconfiguration}
\begin{figure}[htb!]
\includegraphics[scale=0.52]{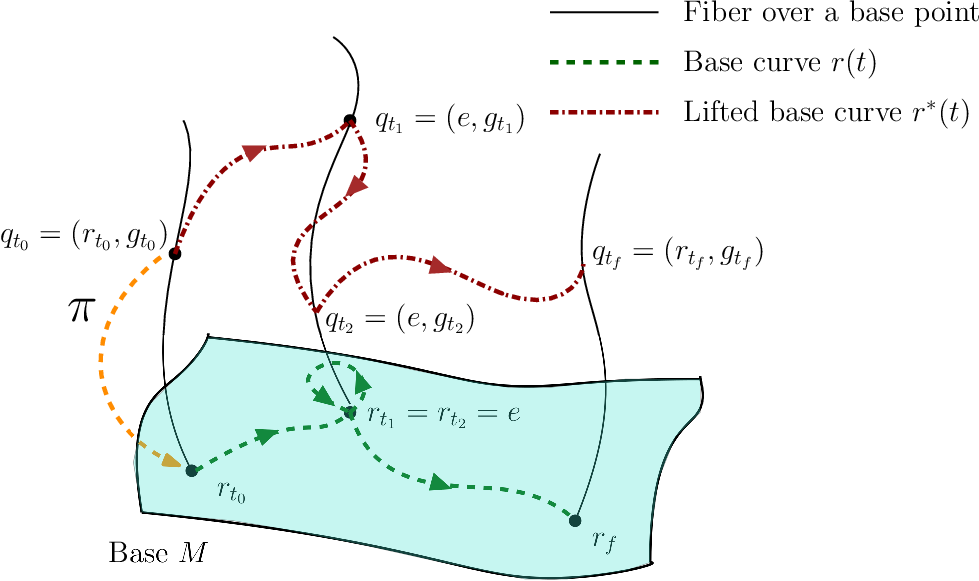}
\caption{Control strategy for point to point reconfiguration}
\label{PointtoPointpathonPFB}
\end{figure}

Given the weak controllability result we now present a strategy for point-to-point reconfiguration of the 3D Purcell's swimmer from any given point $q_{t_0} \in Q$ to any other point $q_{t_f}\in Q$ on its configuration space $Q$. The algorithm for steering from any arbitrary position $q_{t_0} = (r_{t_0}, g_{t_0}) \in Q$ to $q_{t_f} = (r_{t_f}, g_{t_f}) \in Q$ in time $t \in [t_0, t_f]$, with $t_1$ and $t_2$ as intermediate time instants such that $0 \leq t_1 \leq t_2 \leq t_f$ is as follows.
\begin{enumerate}
\item Choose a path $r(t)$ in the base space on time interval $[t_0, t_1]$ such that $q(t_0) = (r_{t_0}, g_{t_0})$ and $q(t_1) = (e, g_{t_1})$, $e$ is the identity of the shape space $SO(3) \times SO(3)$, which also corresponds to the collinear configuration.
\item On the basis of the weak controllability result, starting from $q_{t_1} = (e, g_{t_1})$, we can obtain a control sequence to reach the point $q_{t_2} = (e, g_{t_2})$ such that $g_{t_f}g_{t_2}^{-1}$ is the group displacement in going from the collinear shape $e$ to $r_{t_f}$.
\item We then choose a path $r(t)$ in the base space, giving a shape change from $e$ to $r_{t_f}$, and which results in $g_{t_f}g_{t_2}^{-1}$ as the group displacement in time $[t_2, t_f]$. i.e. $r(t_f) = r_{t_f}$ and $g(t_f) = g_{t_f}$
\end{enumerate}
This sequence of $4$ motions gives a way for point-to-point reconfiguration of the 3D Purcell's swimmer. Figure \ref{PointtoPointpathonPFB} shows the schematic of this sequence  $r(t)$ and its horizontal lift $r^*(t)$ on a principal fiber bundle.

\section{Gait design using motion primitives}
A natural question which arises after showing the controllability of a system is how to synthesize the controls to achieve a given control objective. In our work we seek to control the motion of the swimmer by applying primitive controls. In this section we define the notion of primitive controls and prove the existence of motion primitives based control sequence for the 3D Purcell's swimmer. We then synthesize the control sequences based on the existing work for the planar version of the swimmer. These sequences can be used to achieve the arbitrary group displacements $g_{t_2}g_{t_1}^{-1}$, corresponding to the step $2$ of the algorithm given in the previous section, while remaining at the collinear configuration at the initial and final times.
\subsection{Existence of motion primitive controls}
For $m$ dimensional control inputs, let $U = \{b_1, \cdots, b_m, -b_1, \cdots, -b_m\}$, where $b_i$'s are the unit inputs on $i$'th control. We denote by $\mathcal{U}_{prim}^m$ the collection of piecewise constant $U$-valued controls. Then the motion planning problem using primitives is the one which selects control actions from the set $\mathcal{U}_{prim}^m$. In this case, a controlled trajectory will be a composition of the integral curves of the vector fields $X_1, \cdots, X_m$. We denote by $\Phi_t^X (q)$ the integral curve or the flow of the vector field $X$ for time duration $t$ starting from $q \in Q$. We make use of the following theorem for motion planning using primitives \cite{bullo2004geometric}.
\begin{theorem}
Let $(Q, \mathcal{V}=\{X_1, \cdots , X_m \}, \mathbb{R}^m)$ be a $C^{\infty}$ - driftless system with the vector fields $X_1, \cdots, X_m$ complete. Suppose that $Lie^{(\infty)}(V) = TQ$. If $Q$ is  connected, then, for each $q_0, q_1 \in Q$, there exist $k \in \mathbb{N}$, $t_1, \cdots, t_k \in \mathbb{R}$ and $a_1, \cdots, a_k \in \{1, \cdots, m\}$ such that 
\begin{equation}\label{primitives}
q_1 = \Phi_{t_k}^{X_{a_k}} \cdots \Phi_{t_1}^{X_{a_1}} (q_0)
\end{equation}
\end{theorem}
Both $SO(3)$ and $SE(3)$ are connected manifolds. Further, the product of connected manifolds is connected, hence the Purcell's swimmer's configuration space $SO(3) \times SO(3) \times SE(3)$ is a connected manifold. A vector field on $Q$ is complete if every Cauchy sequence converges in $Q$. We show in appendix \ref{C} that for the 3D Purcell's swimmer, the control vector fields $g_1^1 \cdots g_3^1, g_1^2 \cdots g_3^2 $ in equation (\ref{pure_kinematic_CVFs}) are complete. Hence, using theorem 1, we conclude that there exist motion primitives for our system satisfying (\ref{primitives}). 

This implies that given the weak controllability result and the existence of motion primitives for just the group kinematics in the system given by equation (\ref{pure_kinematic2}), we can find the control sequence which will generate motion in all the 6 basis direction of $\mathfrak{se}(3)$, the Lie algebra of the structure group of the 3D Purcell's swimmer. 

\subsection{Motion primitive controls for the planar Purcell's swimmer}
The control sequence for the planar Purcell's swimmer whose group space is $SE(2)$ has been synthesized for point to point maneuver using symmetry arguments in \cite{gutman2016symmetries}. This problem is looked at from the motion primitives perspective in \cite{kadam2017trajectory}. These 2 references give a set of sequences of control inputs for the planar Purcell's swimmer giving motion along the $\frac{\partial}{\partial x} , \: \frac{\partial}{\partial y} $ and  $\frac{\partial}{\partial \theta}$ directions for the structure group $SE(2)$ of the planar Purcell's swimmer. For this swimmer the actuation is through the limb angles $\psi_1, \psi_2$ corresponding to the angular positions of links 1 and 2 in a plane, such that $\psi_1=0, \psi_2=0$ in the configuration where all the 3 links are collinear.

\begin{figure}[htb!]
\begin{center}
\includegraphics[scale=.27]{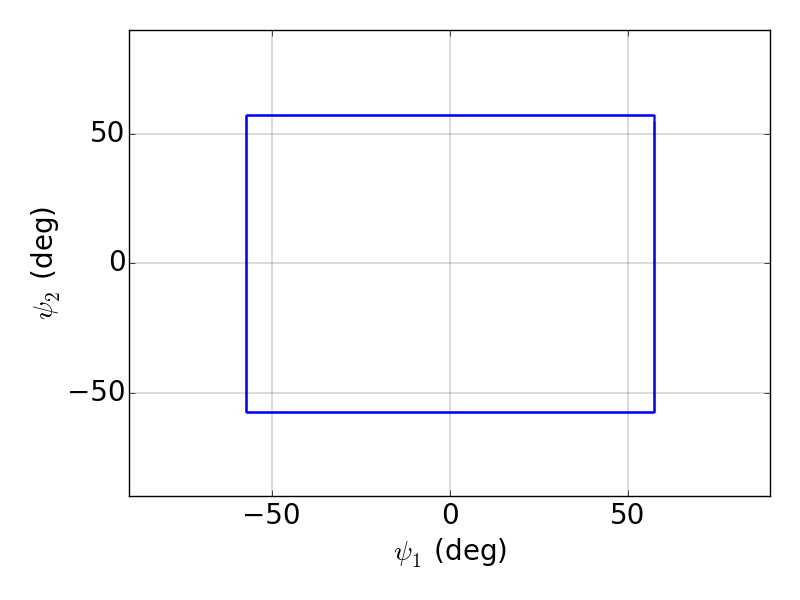}
\caption{X motion}
\label{Planar_Purcell_X_motion}
\end{center}
\end{figure}

\begin{figure}[htb!]
\begin{center}
\includegraphics[scale=.36]{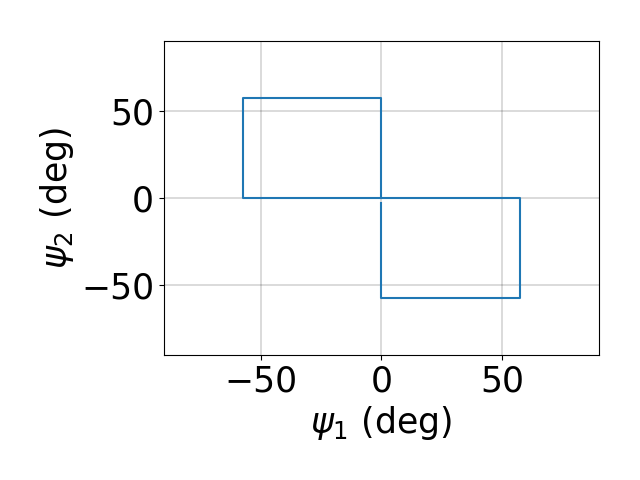}
\caption{Y motion}
\label{Planar_Purcell_Y_motion}
\end{center}
\end{figure}

\begin{figure}[htb!]
\begin{center}
\includegraphics[scale=.36]{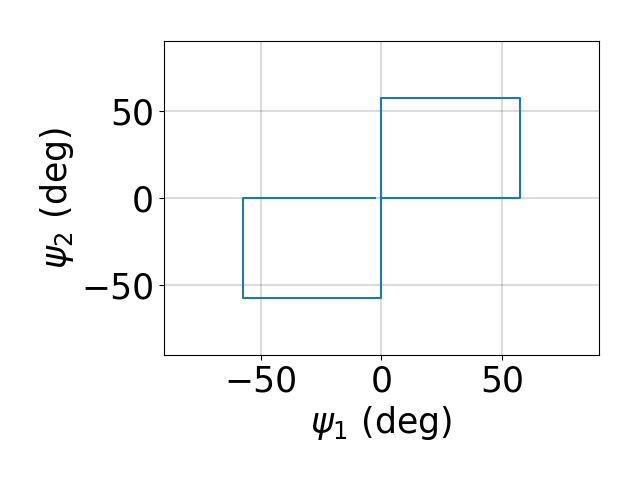}
\caption{Rotation}
\label{Planar_Purcell_angular_motion}
\end{center}
\end{figure}

For the locomotion systems evolving on principal fiber bundle, the base variables are, in general, the control variables. Hence, the control sequences can be effectively shown using gait on the base space, which is any closed path in the base manifold. Figures \ref{Planar_Purcell_X_motion} to \ref{Planar_Purcell_angular_motion} show the gaits for the planar Purcell's swimmer for group motions along the $\frac{\partial}{\partial x} , \: \frac{\partial}{\partial y} $ and  $\frac{\partial}{\partial \theta}$, referred from \cite{gutman2016symmetries}. For all the gaits in the subsequent part of the paper, we consider limb motion amplitude of 1 radian = 57.29 deg.


\subsection{Motion primitive controls for the 3D Purcell's swimmer}
For the 3D Purcell's swimmer, we showed that both $\omega_1$ and $\omega_2$ evolve on the Lie algebra $\mathfrak{so}(3)$ of $SO(3)$. We parametrize orientation $R \in SO(3)$ of the body frame fixed to the base link, shown in figure \ref{fig:3d_coord_free}, with respect to the reference frame by Euler angle sequence $\Psi,\, \theta,\, \phi$, such that $R = R_x^{\phi}R_y^{\theta}R_z^{\psi}$.
\begin{align*}
R_z^{\psi}&=\begin{bmatrix}
    \cos \psi   		& \sin \psi 		& 0\\
    -\sin \psi 	    	& \cos \phi 		& 0 \\
    0					& 0 				& 1
\end{bmatrix}, \: R_y^{\theta}=\begin{bmatrix}
    \cos \theta   	& 0 		& -\sin \theta\\
    0 	    			& 1 		& 0 \\
    \sin \theta		& 0 		& \cos \theta
\end{bmatrix}, \\
R_x^{\phi}&=\begin{bmatrix}
    1   	& 0 				& 0 \\
    0 	    & \cos \phi 		& \sin \phi \\
    0		& -\sin \phi 		& \cos \phi
\end{bmatrix}
\end{align*}
Similarly, the orientations $R_1, R_2$ of the outer links 1 and 2 with respect to the base frame are parametrized using angles $\Psi_1, \theta_1, \phi_1$ for the link 1 and $\Psi_2, \theta_2, \phi_2$ for link 2, such that $R_1 = R_x^{\phi_1}R_y^{\theta_1}R_z^{\psi_1}, \: R_2 = R_x^{\phi_2}R_y^{\theta_2}R_z^{\psi_2}$.

\begin{figure}[htb!]
\begin{center}
\includegraphics[scale=.38]{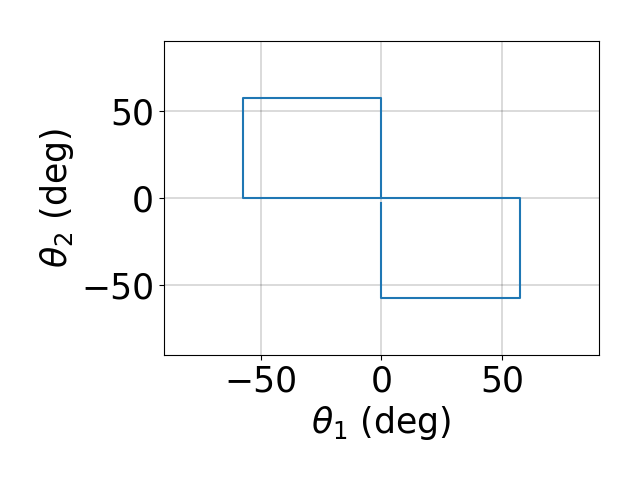}
\caption{Z motion}
\label{Z motion}
  \end{center}
\end{figure}

\begin{figure}[htb!]
\begin{center}
\includegraphics[scale=.38]{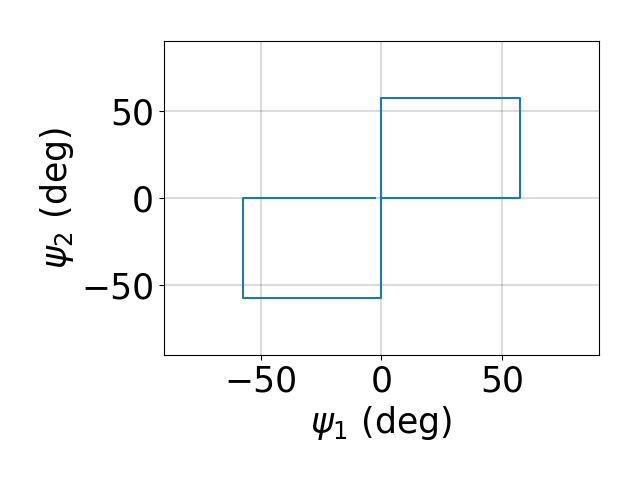}
\caption{Rotation about $z^0$}
\label{GaitAngularmotioninX-Y plane}
\end{center}
\end{figure}

\begin{figure}[htb!]
\begin{center}
\includegraphics[scale=.38]{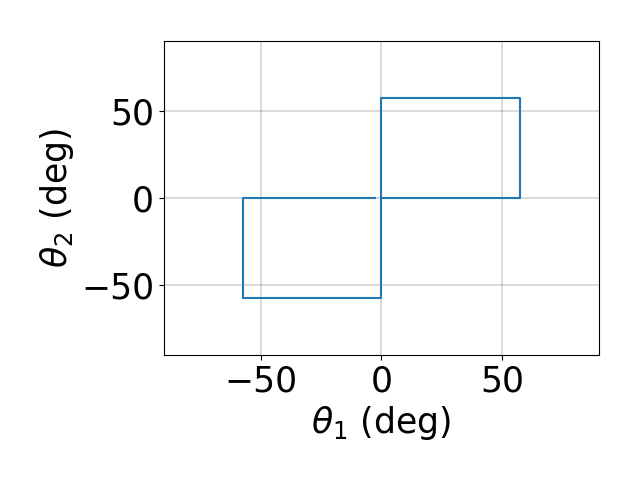}
\caption{Rotation about $y^0$}
\label{GaitAngularmotioninX-Zplane}
  \end{center}
\end{figure}

The motion primitives for the planar Purcell's swimmer presented in the previous section can now be extended to the 3D  swimmer to obtain the control sequences giving motions along the $\frac{\partial}{\partial x} , \: \frac{\partial}{\partial y} , \: \frac{\partial}{\partial z}$ and  $\frac{\partial}{\partial \psi},\: \frac{\partial}{\partial \theta},\: \frac{\partial}{\partial \phi}$ directions for the structure group $SE(3)$ of the 3D Purcell's swimmer. This is based on the fact that in its collinear configuration the actuations of the 3D swimmer can be looked at as actuations of 2 planar Purcell's swimmer moving in 2 orthogonal planes as we did in the controllability analysis. The resulting primitive control sequences for the translational motion of the base link in $X$ and $Y$ directions and the rotational motion remain the same as shown in figures \ref{Planar_Purcell_X_motion}, \ref{Planar_Purcell_Y_motion} and \ref{Planar_Purcell_angular_motion}, respectively. The control sequences for the remaining motion are shown in figures \ref{Z motion} to \ref{GaitAngularmotioninX-Zplane} as gaits on a part of the base manifold $SO(3) \times SO(3)$, parametrized by the Euler angles. 


%

These control sequences can be written in terms of the flow of the control vector fields in equation (\ref{pure_kinematic_CVFs}). The following sequence of actuation leads to the pure translational motion of the base link

\begin{align*}
\text{Motion along $\: \frac{\partial}{\partial x}$ : } & \Phi^{-g_3^2\omega_2^3}_{t} \circ \Phi^{-g_3^1\omega_1^3}_{t} \circ \Phi^{g_3^2\omega_2^3}_{t} \circ \Phi^{g_3^1\omega_1^3}_{t} (\textbf{e})\\
\text{Motion along $\: \frac{\partial}{\partial y}$ : } & \Phi^{g_3^2\omega_2^3}_{t} \circ \Phi^{-g_3^1\omega_1^3}_{t} \circ \Phi^{-g_3^2\omega_2^3}_{t} \circ \Phi^{g_3^1\omega_1^3}_{t} \\
&\circ \Phi^{g_3^1\omega_1^3}_{t} \circ \Phi^{-g_3^2\omega_2^3}_{t} \circ \Phi^{-g_3^1\omega_1^3}_{t} \circ \Phi^{g_3^2\omega_2^3}_{t} (\textbf{e}) \\
\text{Motion along $\: \frac{\partial}{\partial z}$ : } &  \Phi^{g_2^2\omega_2^2}_{t} \circ \Phi^{-g_2^1\omega_1^2}_{t} \circ \Phi^{-g_2^2\omega_2^2}_{t} \circ \Phi^{g_2^1\omega_1^2}_{t} \\
& \circ \Phi^{g_2^1\omega_1^2}_{t} \circ \Phi^{-g_2^2\omega_2^2}_{t} \circ \Phi^{-g_2^1\omega_1^2}_{t} \circ \Phi^{g_2^2\omega_2^2}_{t} (\textbf{e})
\end{align*}
where $\textbf{e}$ is the identity element of the full configuration space $Q$ which denotes the collinear configuration of the swimmer and the coordinate frame associated with the base link coinciding the inertial frame. The control sequences for the pure rotational motions of the base link are as follows

\begin{align*}
\text{Motion along $\frac{\partial}{\partial {\psi}}$ : } & \Phi^{g_3^1\omega_1^3}_{t} \circ \Phi^{g_3^2\omega_2^3}_{t} \circ \Phi^{-g_3^1\omega_1^3}_{t} \circ \Phi^{-g_3^2\omega_2^3}_{t} \\
& \circ \Phi^{-g_3^2\omega_2^3}_{t} \circ \Phi^{-g_3^1\omega_1^3}_{t} \circ \Phi^{g_3^2\omega_2^3}_{t} \circ \Phi^{g_3^1\omega_1^3}_{t} (\textbf{e}) \\
\text{Motion along $\frac{\partial}{\partial {\theta}}$ : } & \Phi^{g_2^1\omega_1^2}_{t} \circ \Phi^{g_2^2\omega_2^2}_{t} \circ \Phi^{-g_2^1\omega_1^2}_{t} \circ \Phi^{-g_2^2\omega_2^2}_{t} \\
&\circ \Phi^{-g_2^2\omega_2^2}_{t} \circ \Phi^{-g_2^1\omega_1^2}_{t} \circ \Phi^{g_2^2\omega_2^2}_{t} \circ \Phi^{g_2^1\omega_1^2}_{t} (\textbf{e}) \\
\text{Motion along $\frac{\partial}{\partial {\phi}}$ : } & \Phi^{g_1^1\omega_1^1}_{t} (e) \text{ or } \Phi^{g_1^2\omega_2^1}_{t} (\textbf{e})
\end{align*}

Thus we have the expression for synthesizing the sequence of primitive control and the corresponding actuation-durations to achieve a velocity in a given group direction. 

\subsection{Simulation results}
\begin{figure}[htb!]
\begin{center}
\includegraphics[width=0.35\textwidth]{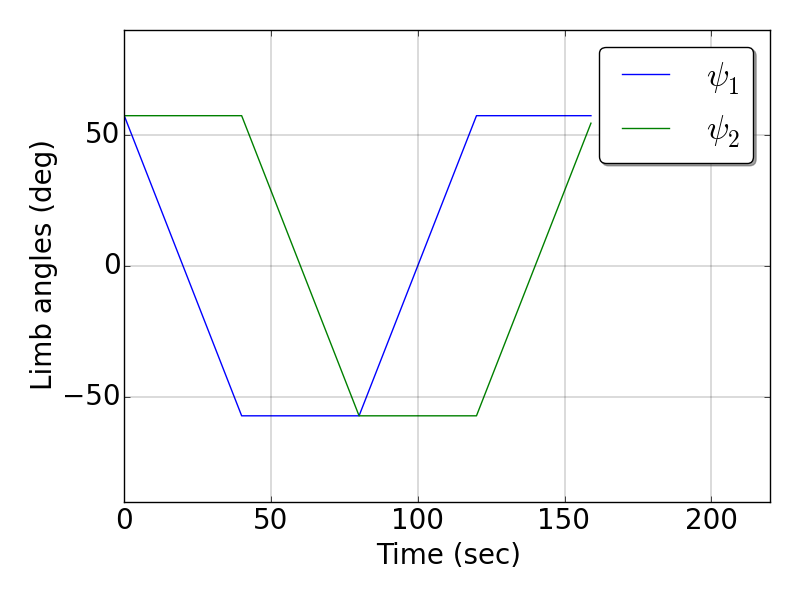}
\caption{Limb angles vs time for X motion}
\label{LimbanglesvstimeforXmotion}
\end{center}
\end{figure}

\begin{figure}[htb!]
\begin{center}
\includegraphics[width=0.35\textwidth]{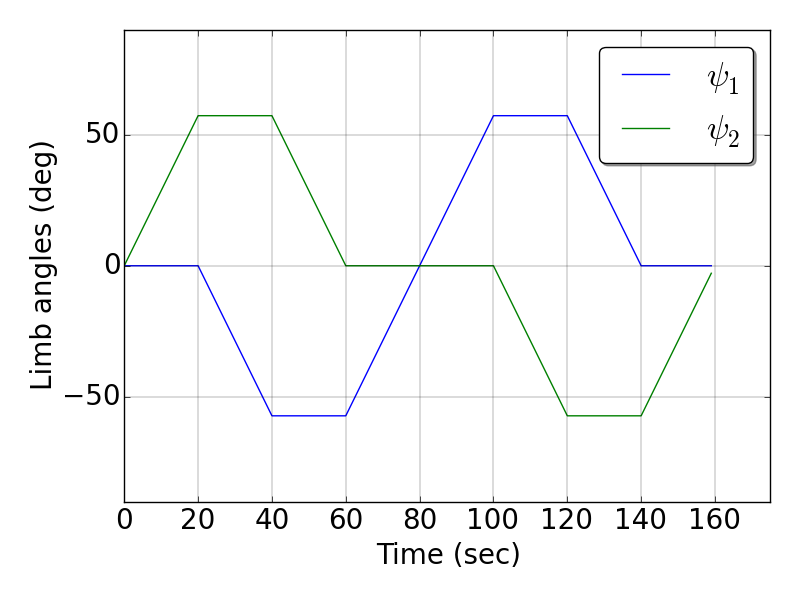}
\caption{Limb angles vs time for Y motion}
\label{LimbanglesvstimeforYmotion}
  \end{center}
\end{figure}

\begin{figure}[htb!]
\begin{center}
\includegraphics[width=0.35\textwidth]{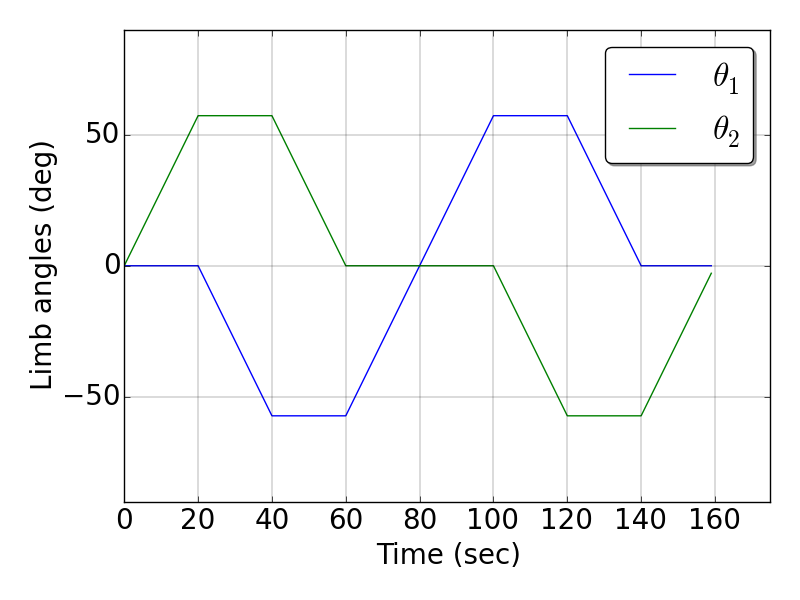}
\caption{Limb angles vs time for Z motion}
\label{LimbanglesvstimeforZmotion}
  \end{center}
\end{figure}

\begin{figure}[htb!]
\begin{center}
\includegraphics[width=0.35\textwidth]{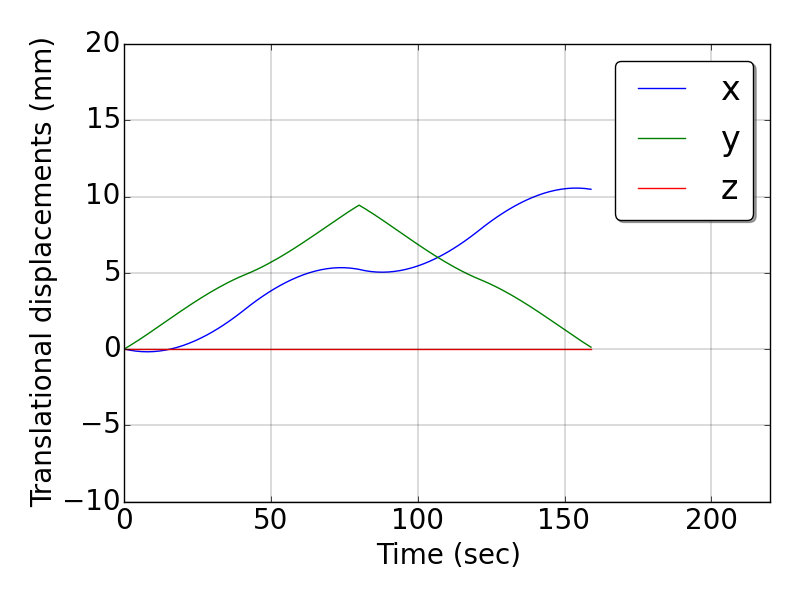}
\caption{Translational positions vs time for X motion}
\label{TranslationalpositionsvstimeforXmotion}
\end{center}
\end{figure}

\begin{figure}[htb!]
\begin{center}
\includegraphics[width=0.35\textwidth]{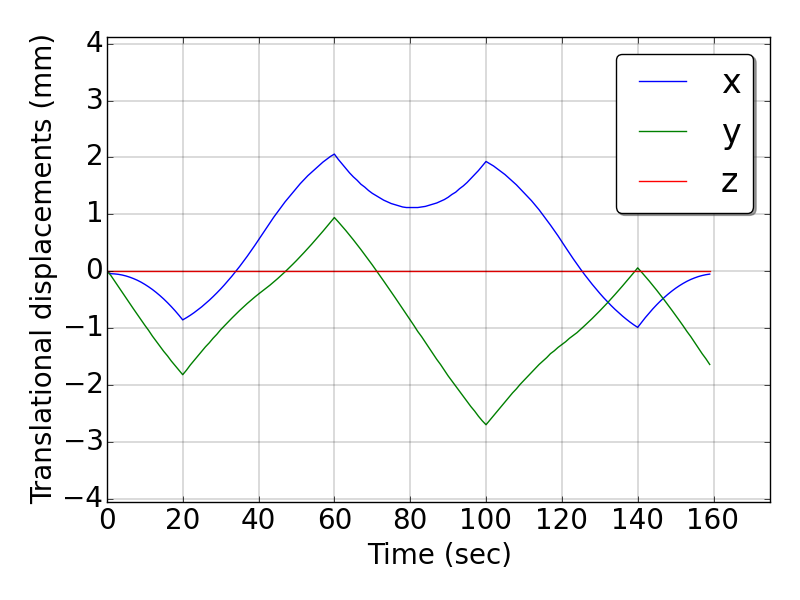}
\caption{Translational positions vs time for Y motion}
\label{TranslationalpositionsvstimeforYmotion}
  \end{center}
\end{figure}

\begin{figure}[htb!]
\begin{center}
\includegraphics[width=0.35\textwidth]{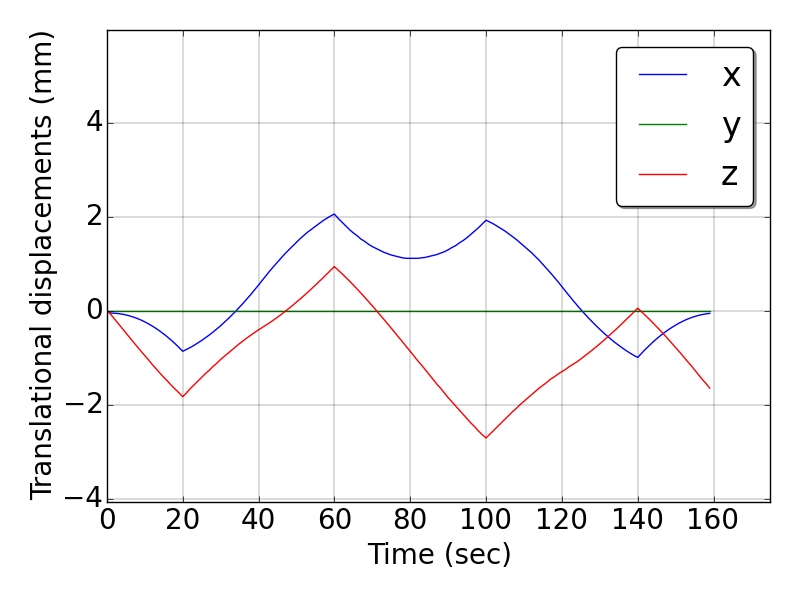}
\caption{Translational positions vs time for Z motion}
\label{TranslationalpositionsvstimeforZmotion}
  \end{center}
\end{figure}

\begin{figure}[htb!]
\begin{center}
\includegraphics[width=0.35\textwidth]{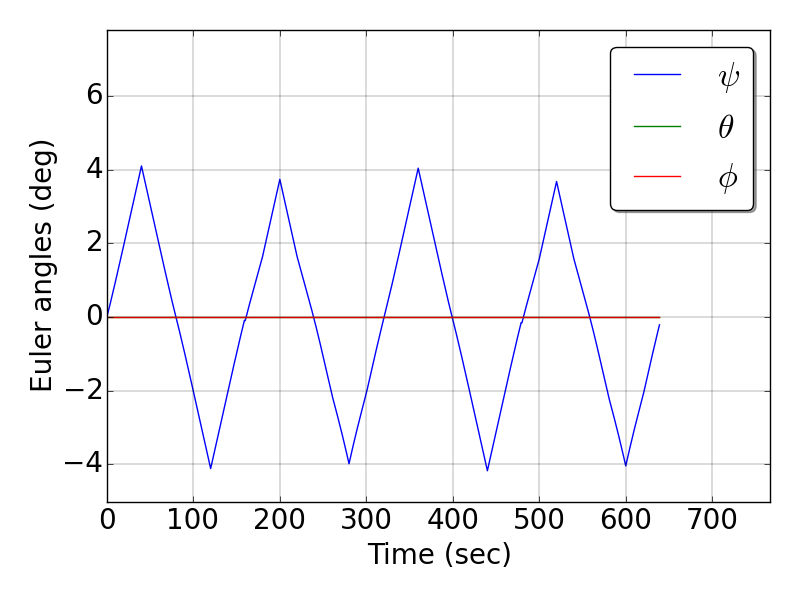}
\caption{Euler angles vs time for Z motion}
\label{EuleranglesvstimeforXmotion}
\end{center}
\end{figure}

\begin{figure}[htb!]
\begin{center}
\includegraphics[width=0.35\textwidth]{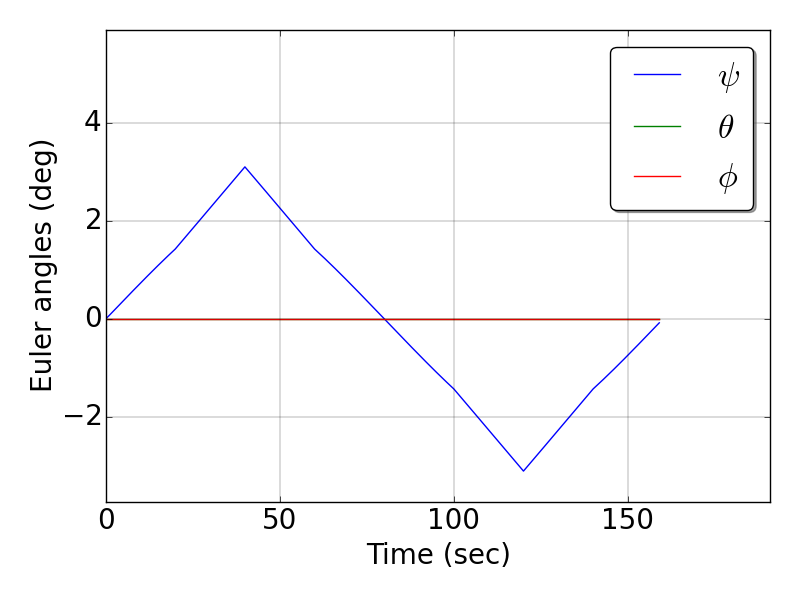}
\caption{Euler angles vs time for Y motion}
\label{EuleranglesvstimeforYmotion}
  \end{center}
\end{figure}

\begin{figure}[htb!]
\begin{center}
\includegraphics[width=0.35\textwidth]{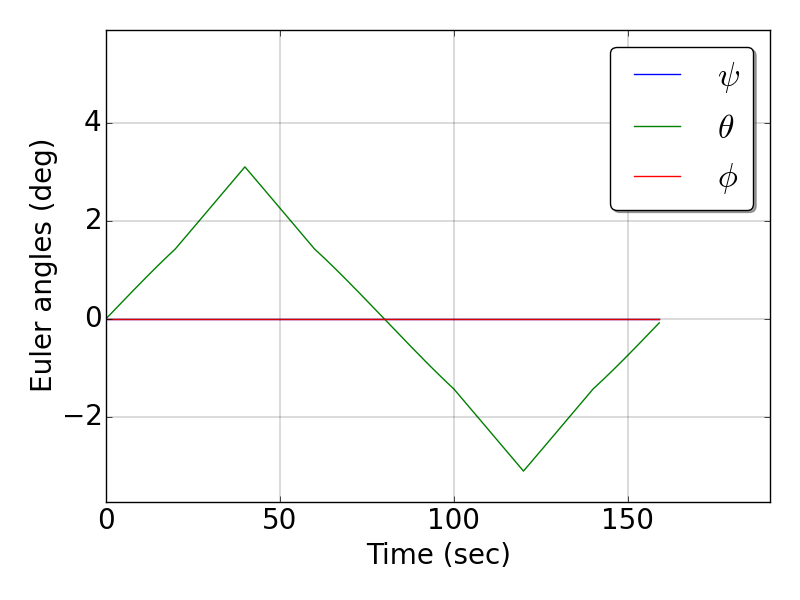}
\caption{Euler angles vs time for Z motion}
\label{EuleranglesvstimeforZmotion}
  \end{center}
\end{figure}

\begin{figure}[htb!]
\begin{center}
\includegraphics[width=0.35\textwidth]{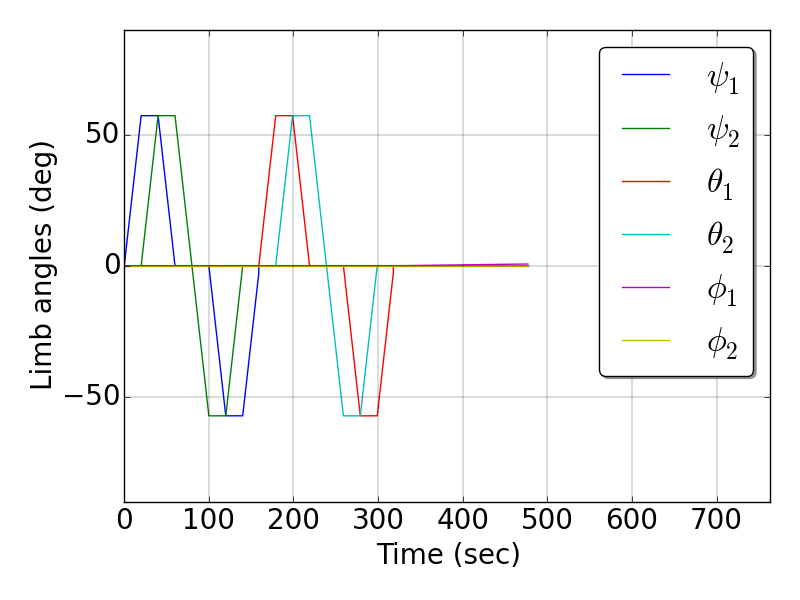}
\caption{Limb angles vs time for rotational maneuver}
\label{LimbanglesvstimeforRotation}
\end{center}
\end{figure}

\begin{figure}[htb!]
\begin{center}
\includegraphics[width=0.35\textwidth]{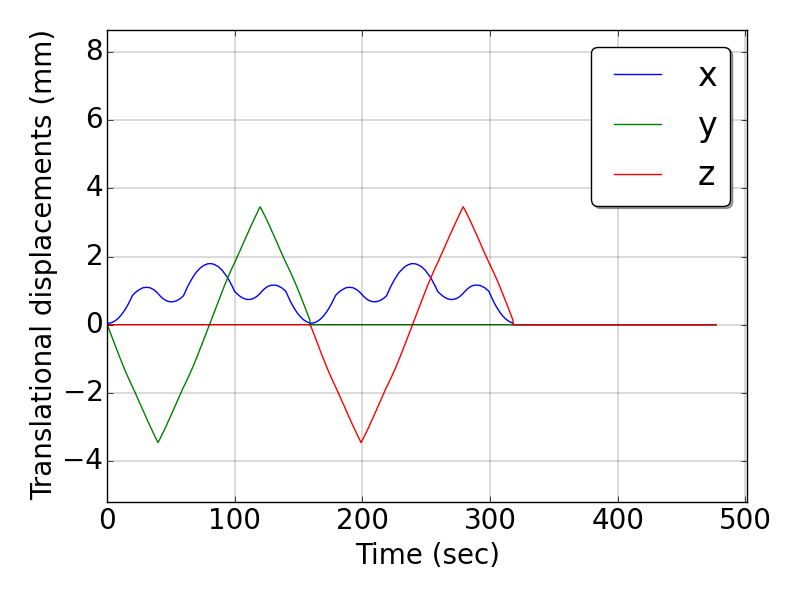}
\caption{Translational positions for rotational maneuver}
\label{TranslationalpositionvstimeforRotation}
  \end{center}
\end{figure}

\begin{figure}[htb!]
\begin{center}
\includegraphics[width=0.35\textwidth]{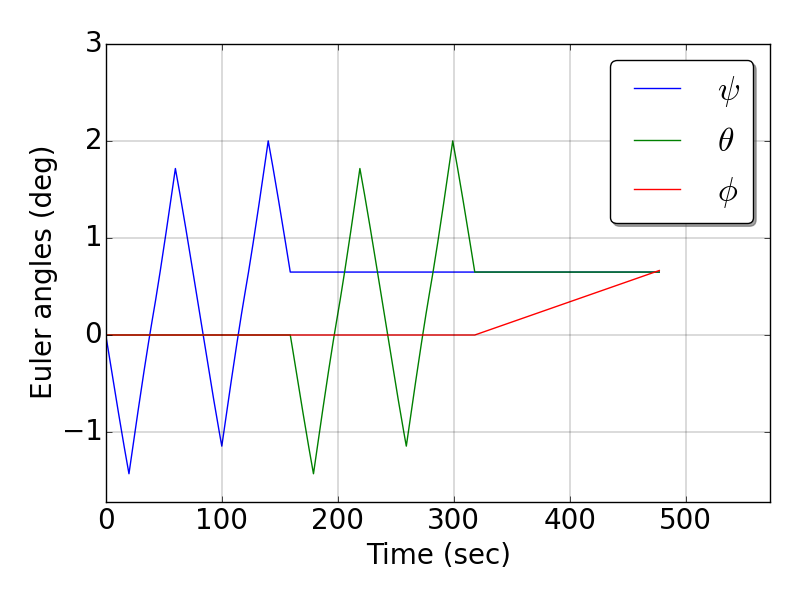}
\caption{Euler angles vs time for rotational maneuver}
\label{EuleranglesvstimeforRotation}
  \end{center}
\end{figure}

In this section, we present results of the simulation we performed for the 3D Purcell's swimmer to achieve the motions along the six basis elements of $se(3)$ using the motion primitives based control sequences presented in the previous subsection. The simulation was carried out with link length $L = 100mm$, slenderness ratio $0.01$, the fluid was considered to be glycerin which has dynamic viscosity $\nu$ of $0.95 PaS$. The viscous drag coefficient for the simulation was computed by the method described in section III. Figures \ref{LimbanglesvstimeforXmotion} to \ref{EuleranglesvstimeforRotation} show the results of the simulation depicting the variations in limb angles of the outer links and translation position and Euler angles of the base link for translational motion along the x, y and z directions of the reference frame.

Figures \ref{LimbanglesvstimeforRotation} to \ref{EuleranglesvstimeforRotation} show the variation of limb Euler angles, translation displacements and the base link's Euler angle variation against time.

\section{Conclusion}
This paper introduces a generalized version of the Purcell's swimmer which performs 3 dimensional motion. We highlight the topology of a trivial principal fiber bundle that the configuration space of this swimmer inherits, and derive a principal kinematic form of equations by utilizing the properties at low Reynold's number regime. Matrix Lie groups are used in order to get a coordinate free expression for the local connection form. The local weak controllability of the 3D swimmer is proved by invoking the controllability results of the planar Purcell's swimmer. The decomposition of the limb motion of the 3D swimmer in 2 orthogonal planes has been utilized to understand the achievable group motions of the swimmer's base link through the weak controllability analysis. An algorithm for point-to-point reconfiguration of the swimmer is presented by making use of the weak controllability result. We then showed the existence of motion primitives based control sequence for the swimmer. A control sequence to generate translational and rotation motions for the base link of the swimmer was synthesized followed by a numerical simulation. An interesting avenue of future work is to explore optimal control and motion planning strategies for the 3D swimmer presented here. We hope to work on these issues in the future.

\section*{Acknowledgment}
A part of this work was accomplished by the second author during his tenure as a Pratt and Whitney Chair Professor in the Department of Aerospace Engineering, Indian Institute of Sciences, Bengaluru, India.

\section{Appendix A}\label{B}
In this section we describe the notions of horizontal lift of a base curve and local weak controllability for systems on principal fiber bundle, for which a point in the configuration space is written as $q = (r,g) \in M \times G = Q$. For a base curve $r(t) \in M$, the horizontal lift $r^*(t) \in Q$ is a curve which projects to $x(t)$ under the projection map defining the principal fiber bundle and the components of its tangent vectors $\dot{r}^*(t) \in T_qQ$ satisfy the reconstruction equation (\ref{Final_kinematic_model}). 

Such a system is said to be weakly controllable if, for any initial position $g_0 \in G$, and final position $g_f \in G$, and initial shape $r_0 \in M$, there exists a time $T > 0$ and a curve in the base space $r(t)$ satisfying $r(0) = r_0$ such that the horizontal lift of $r(t)$ passing through $(r_0, g_0)$  satisfies $r^*(0) = q_0$ and $r^*(T) = (r(T), g_f)$ \cite{kelly1995geometric}.

To find the analytical condition of the weak controllability, we denote by $DA(x)$, the curvature of the local connection form $A$ at $r \in M$, which is a Lie algebra valued differential 2-form. We denote by $L_Z DA(r)$ the Lie derivative of the curvature form with respect to the vector field on the base space $M$. Then, given the following vector spaces,
\begin{align*}
\mathfrak{h}_1\:&=\: span \{A(r)(X) : X \in T_r M\},\\
\mathfrak{h}_2\:&=\: span \{DA(r)(X,Y) : X, Y \in T_r M\}, \\
\mathfrak{h}_3\:&=\: span \{L_Z DA(r)(X,Y) - [A(r)(Z),DA(r)(X,Y)], \\ & \qquad [DA(r)(X,Y),DA(x)(W,Z)] : W, X, Y, Z \in T_rM\} \\
\vdots \\
\mathfrak{h}_k\:&=\: span \{L_X \xi - [A(r)(X),\xi],[\eta,\xi] :
X \in T_r M, \\
& \:\: \qquad\qquad \xi \in \mathfrak{h}_{k-1}, \eta \in \mathfrak{h}_2 \:\oplus\: \cdots \:\oplus\:\mathfrak{h}_{k-1}\}
\end{align*}
a system defined on a trivial principal bundle $Q$ is locally weakly controllable near $q \in Q$ if and only if the space of the Lie algebra $\mathfrak{g}$ of the structure group $G$ is spanned by the vector fields $\mathfrak{h}_1, \mathfrak{h}_2, \cdots$ as follows
\begin{equation}\nonumber
\mathfrak{g}\:=\:\mathfrak{h}_1\:\oplus\:\mathfrak{h}_2\:\oplus \cdots
\end{equation}

\section{}\label{A}
Hat map, defined as $\times : \mathbb{R}^n \mapsto Skew\, (\mathbb{R}^{n \times n})$ is used on $3$ dimensional vectors in this paper. The expression takes the following form -
\begin{equation}\nonumber
\begin{bmatrix}
a \\
b \\
c
\end{bmatrix}^{\times} = \begin{bmatrix}
0   & -c  &b \\
c   & 0   &-a \\
-b  & a   &0
\end{bmatrix}
\end{equation}
The hat map is useful in writing a cross product of 2 vectors $x, y$ compactly as a matrix multiplication $x \times y = x^{\times}y = -y^{\times} x$

\section{}\label{C}
\begin{theorem}
The control vector fields $g_1^1 \cdots g_3^1, g_1^2 \cdots g_3^2 $ for the 3D Purcell's swimmer in the equation (\ref{pure_kinematic_CVFs}) are complete.
\end{theorem}
\begin{proof}
\begin{itemize}
\item The domain of all of the control vector fields $g_1^1 \cdots g_3^1, g_1^2 \cdots g_3^2 $ is $SO(3) \times SO(3)$. We observe that both $g_1$ and $g_2$ never attend zero value in their domain, since there is a constant value $1$ appearing in one of the components of both $g_1, g_2$. Hence, the support of $g_1$ and $g_2$ is the entire domain $SO(3) \times SO(3)$
\item $SO(3)$ is compact because its a closed subgroup of compact group $O(3)$
\item Cartesian product of $2$ compact manifolds is compact, hence $SO(3) \times SO(3)$ is also compact \cite{sutherland2009introduction}. Thus, the support of control vector fields $g_1^1 \cdots g_3^1, g_1^2 \cdots g_3^2 $ is compact
\item A vector field is complete if its support is compact \cite{Cryster12}. Hence, $g_1^1 \cdots g_3^1, g_1^2 \cdots g_3^2 $ are complete vector fields.
\end{itemize}
\end{proof}

\bibliographystyle{ieeetran}
\bibliography{bib1}

%

\end{document}